\documentclass{article}

\PassOptionsToPackage{numbers, sort, compress}{natbib}
\usepackage[final]{neurips_2023}

\usepackage[utf8]{inputenc} 
\usepackage[T1]{fontenc}    
\usepackage{hyperref}       
\usepackage{url}            
\usepackage{booktabs}       
\usepackage{amsfonts}       
\usepackage{nicefrac}       
\usepackage{microtype}      
\usepackage{xcolor}         
\usepackage{amsmath, amssymb, amsthm}
\usepackage{dsfont}
\usepackage{comment}
\usepackage{graphicx}
\usepackage{floatrow}
\usepackage{rotating}
\usepackage{enumitem}
\usepackage{multirow}

\usepackage{wrapfig}
\usepackage{caption}
\usepackage{subcaption}
\usepackage{caption}
\usepackage{array}
\usepackage{nicematrix}
\usepackage{bm}
\usepackage{colortbl}

\usepackage[ruled,vlined]{algorithm2e}

\def\*#1{\mathbf{#1}}

\makeatletter
\DeclareRobustCommand\onedot{\futurelet\@let@token\@onedot}
\def\@onedot{\ifx\@let@token.\else.\null\fi\xspace}

\def\eg{\emph{e.g}\onedot} 
\def\ie{\emph{i.e}\onedot}

\def\wrt{\emph{w.r.t}\onedot} 

\makeatother

\title{GAIA: Delving into Gradient-based Attribution Abnormality for Out-of-distribution Detection}

\author{%
  Jinggang Chen$^{\dag}$\footnotemark[1] , Junjie Li$^{\dag*}$, Xiaoyang Qu$^{\ddag}$\footnotemark[4] , Jianzong Wang$^{\ddag}$\footnotemark[4] , Jiguang Wan\footnotemark[2] , Jing Xiao\footnotemark[3] \\
  $^{\dag}$ Huazhong University of Science and Technology, China \\
  $^{\ddag}$ Ping An Technology (Shenzhen) Co., Ltd.\\
  \{\texttt{chen.jinggang98, 2216217669ljj, quxiaoy\}@gmail.com, jzwang@188.com,} \\ \texttt{jgwan@hust.edu.cn, xiaojing661@pingan.com.cn}\\
}

\begin{document}
\renewcommand{\thefootnote}{\fnsymbol{footnote}}

\footnotetext[1]{Equal Contribution.}
\footnotetext[4]{Corresponding Author.}
 


\maketitle

\renewcommand{\thefootnote}{\arabic{footnote}}

\begin{abstract}
Detecting out-of-distribution (OOD) examples is crucial to guarantee the reliability and safety of deep neural networks in real-world settings. In this paper, we offer an innovative perspective on quantifying the disparities between in-distribution (ID) and OOD data---analyzing the uncertainty that arises when models attempt to explain their predictive decisions. This perspective is motivated by our observation that gradient-based attribution methods encounter challenges in assigning feature importance to OOD data, thereby yielding divergent explanation patterns. Consequently, we investigate how attribution gradients lead to uncertain explanation outcomes and introduce two forms of abnormalities for OOD detection: the zero-deflation abnormality and the channel-wise average abnormality. We then propose \textbf{GAIA}, a simple and effective approach that incorporates \textbf{G}radient \textbf{A}bnormality \textbf{I}nspection and \textbf{A}ggregation.  The effectiveness of GAIA is validated on both commonly utilized (CIFAR) and large-scale (ImageNet-1K) benchmarks. Specifically, GAIA reduces the average FPR95 by 23.10\% on CIFAR10 and by 45.41\% on CIFAR100 compared to advanced post-hoc methods.

\end{abstract}

\section{Introduction}
    
Deep neural networks have been extensively applied across various domains, demonstrating remarkable performance. However, when they are deployed in real-world scenarios, particularly in contexts that require high levels of security \cite{huang2020survey, medical, fiance}, an urgent challenge arises. Namely, these models must be able to ensure the reliability of their outcomes, even in the face of out-of-distribution (OOD) inputs from the open world that differ from in-distribution (ID) training data and thus surpass their cognitive capabilities. That underscores the importance of OOD detection, which involves estimating uncertainty from the model to identify the "unknown" samples, serving as an alert mechanism before making predictive decisions.

Recently, a rich line of literature has emerged to address the challenge of OOD detection \cite{baseline_G-ODIN, liu2020energy, 1-D, MaxConfidenceScore, lakshminarayanan2017simple, ODIN, grad_norm, React, OE, Gram}.
Indeed, the majority of previous approaches focus on defining more suitable measures of OOD uncertainty by using model outputs \cite{MaxConfidenceScore, baseline_G-ODIN, lakshminarayanan2017simple, ODIN, liu2020energy} or feature representations \cite{ma_distance, Gram, React,rankfeat}. Despite the above mainstream approaches, estimating uncertainty from gradients is readily implemented with a fixed model and
has received increasing research attention lately. Prior gradient-based OOD detection methods \cite{lee2020gradients, grad_norm, gradient_igoe2022useful} have primarily emphasized utilizing parameter gradients as the measurement, while giving limited attention to the in-depth exploration of gradients related to the inputs (\ie, attribution gradients \cite{attribution_sensitivity}).

\begin{figure}[t]
  \centering
  \includegraphics[width=\textwidth]{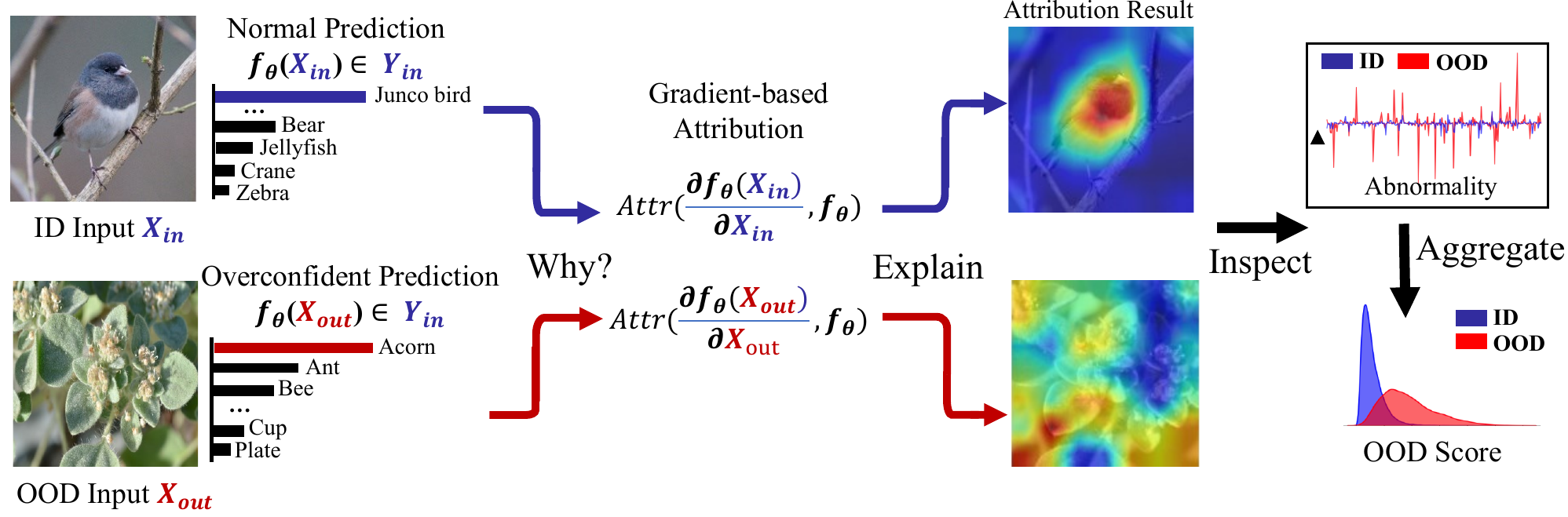}
  \caption{Motivation of our work. Gradient-based attribution algorithms use attribution gradients to explain where models look for predicting final outputs. An intriguing question is: when encountering OOD sample $X_{\text{out}}$ whose label falls outside the in-distribution label space $Y_{\text{in}}$, how does the model interpret its overconfident prediction? In order to unearth uncertainty from the explanatory result, we conduct our research by inspecting the abnormalities in attribution gradients and then aggregate them for OOD detection.}
  \label{fig::motivation_illustration}
\end{figure}
In this paper, we put our eye on a novel and insightful perspective --- let models explain the uncertainty themselves with attribution approaches. Gradient-based attribution algorithms \cite{channel_mean, gradcam++, jiang2021layercam} are ubiquitous for the visual explanation of why the model makes such a decision to the predicted class. An intuition comes up that well-trained networks can clearly attribute the region of target ID objects, but what if they face OOD samples that are totally unknown to them? As shown in Fig. \ref{fig::motivation_illustration}, we observe through the utilization of attribution gradients that the pre-trained model is capable of generating reasonable visual interpretation for the ID input $X_{\text{in}}$ from ImageNet \cite{deng2009imagenet}. However, when attempting to interpret an OOD image $X_{\text{out}}$ from iNaturalist \cite{van2018inaturalist} with a label that does not belong to $Y_{\text{in}}$, it confuses the model, leading to a meaningless attribution result.

Following the observation, we delve into investigating the gradient-based attribution abnormality when inferring OOD examples. Our further study finds that this phenomenon can be caused by the attribution gradient, which is constructed by taking the value of the partial derivative of the target output $S_c(\cdot)$ \wrt one unit $z_i$ of the input variables $\bm{z}$ (\ie, $\frac{\partial S_c(\bm{z})}{\partial z_i}$). To enlarge the discrepancy between ID and OOD without prior knowledge from training data, we introduce the channel-wise average abnormality and the zero-deflation abnormality as two measurements for detecting distributional shifts. Then, we propose our detection framework \textbf{GAIA} with \textbf{G}radient \textbf{A}bnormality \textbf{I}nspection and \textbf{A}ggregation and conduct comprehensive experiments on both CIFAR benchmarks and large-scale ImageNet-1K benchmark to validate the effectiveness of our proposed method. Code is available at \href{https://github.com/JGEthanChen/GAIA-OOD}{https://github.com/JGEthanChen/GAIA-OOD}.

Our key results and contributions are summarized as follows:
\begin{itemize}
    \item We provide insights into the attribution abnormality for OOD detection. Our intuition is that unreliability from visual explanations can be a direct alarm to distinguish OOD examples. Hence, we delve further into the underlying causality of the abnormality. Then, we provide a theoretical explanation for the causes of attribution abnormality.
    
    \item We propose a simple yet effective post-hoc detection framework via \textbf{G}radient \textbf{A}bnormality \textbf{I}nspection and \textbf{A}ggregation (\textbf{GAIA}), which consists of two independent measurements: the Channel-wise Average abnormality (\textbf{GAIA-A}) and the Zero-deflation abnormality (\textbf{GAIA-Z}). Both of them are lightweight and plug-and-play---hyperparameter-free, training-free, with no ID data and outliers required for estimation. 
    
    \item Thorough experiments demonstrate that GAIA surpasses most advanced post-hoc methods on both commonly utilized (CIFAR) and large-scale (ImageNet-1K) benchmarks. GAIA-Z exhibits superior performance on CIFAR benchmarks, reducing the average FPR95 by 23.10\% on CIFAR10 and by 45.41\% on CIFAR100. GAIA-A performs well on the ImageNet-1K benchmark and reduces by 17.28\% compared to the advanced gradient-based detection method GradNorm.

\end{itemize}

\section{Preliminaries}

We consider the general setting of a supervised machine learning problem, where $\mathcal{X}$ denotes the input space and $\mathcal{Y}_{\text{in}}=\{ 1, 2,..., C\}$ denotes the ID label space. Especially, we denote the output score \wrt class $c$ before softmax layer as $S_{c}(\cdot)$.

\textbf{Out-of-distribution detection.} The goal of out-of-distribution (OOD) detection is to distinguish the sample $x_{\text{out}}$ that exhibits substantial deviation from the distribution $\mathcal{X}$. In literature, OOD data originates from an unknown distribution $\mathcal{X}_{\text{out}}$. And the label space of the OOD samples has no intersection with $\mathcal{Y}_{\text{in}}$. This problem can be formulated as a binary classification task using a score function $\Delta(x)$. More specifically, when provided with an input sample $x$, the level-set estimation can be expressed as follows:
\begin{equation}
    \mathcal{G}(x) = \begin{cases}
    \text{out}, &\text{if}\quad\Delta(x) > \gamma\\
    \text{in}, &\text{if}\quad\Delta(x) \leq \gamma\\
    \end{cases}  
\end{equation}
In our work, lower scores correspond to a higher likelihood of classifying the sample $x$ as in-distribution (ID), and $\gamma$ denotes a threshold for separating the ID and OOD data.

\textbf{Gradients from attribution algorithms.} The attribution gradient is first introduced by sensitivity analysis (SA) \cite{attribution_sensitivity} and widely utilized in visual explainability techniques \cite{gradient_input, channel_mean, gradcam++, jiang2021layercam, integrated_gradients}. It refers to the sensitivity of a particular input variable (input or feature unit) \wrt $c$-class predictive output $S_{c}(\cdot)$.
Denotes $k$-th channel feature map at layer $l$ as $\bm{A}^{kl}\in\mathbb{R}^{W \times H}$. The attribution gradient of one feature unit $A^{kl}_{ij}$ is computed by:
\begin{equation}
    \text{Grad}_{ij} =  \frac{\partial S_c(\bm{A}^{kl})}{\partial A^{kl}_{ij}}  
\end{equation}
It is unrelated to the gradients commonly associated with the typical understanding of network optimization (\ie, gradients of the parameters). In most attribution algorithms, the attribution gradient is used for quantifying the contribution of each feature unit to the model's prediction.

\section{Investigating Attribution Abnormality for Out-of-distribution Detection}

\begin{figure}[b]
  \centering
  \includegraphics[width=0.9\textwidth]{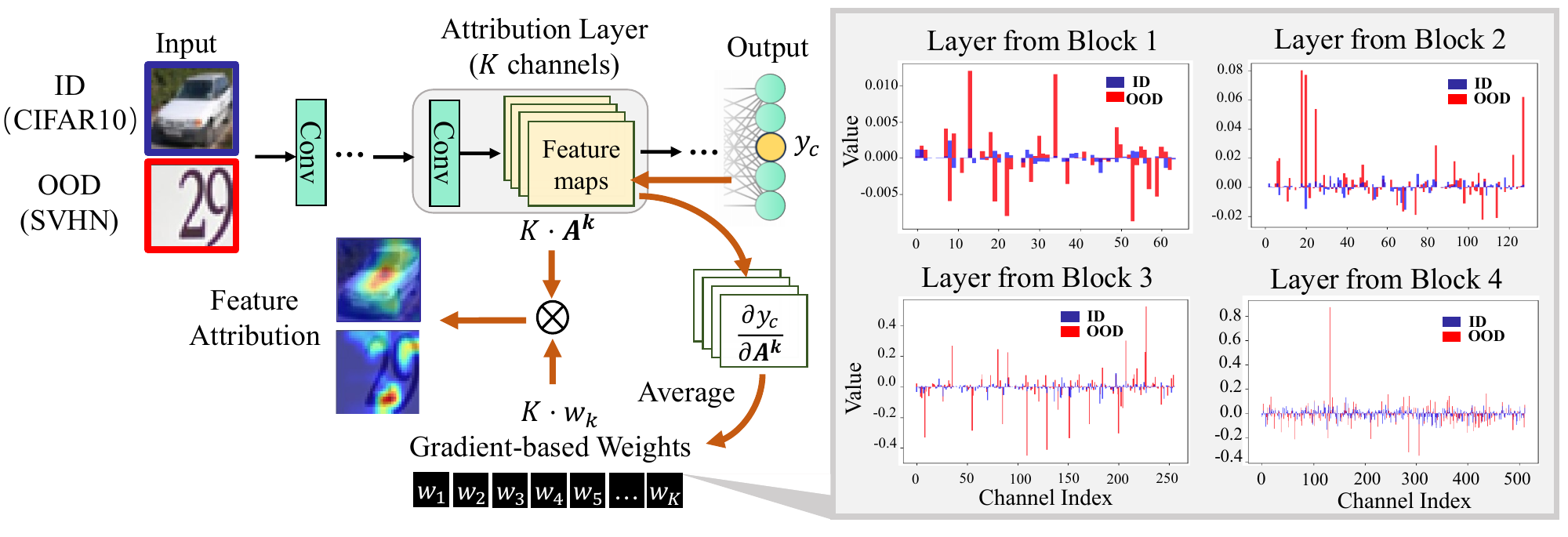}
  \caption{Demonstration of the attribution abnormality from gradient-based weights. The toy experiment is conducted on ResNet34 with four blocks trained on CIFAR10. We select four attribution layers from different blocks and calculate the average attribution gradients for each channel.}
  \label{fig::demo_avg}
\end{figure}

In this section, we aim to investigate how attribution gradients can lead to abnormality when explaining OOD examples. We also attempt to provide a unified theoretical analysis.

\textbf{Channel-wise average abnormality.} We first focus on the abnormality in the Gradient-based Class Activation Mapping (GradCAM) algorithm \cite{channel_mean}, which is one of the most widely applied attribution strategies. Its paradigm is to channel-wise sum up feature maps for a saliency map $\bm{M}\in\mathbb{R}^{W \times H}$. Here we denote feature maps $\bm{A}\in\mathbb{R}^{K\times W \times H}$ with $K$-channels in the convolutional layer as the input variables. The attribution $\alpha_{ij}$ of each unit $M_{ij}$ can be formulated as follows:
\begin{equation}
    \alpha_{ij} = \text{ReLU}(\sum_{k=1}^K w_k A^k_{ij}), \quad\text{where} \quad w_k = \frac{\partial g(\bm{A})}{\partial A^k_{ij}} = \frac{1}{W \times H} \sum^{W}_{i=1} \sum^{H}_{j=1} \frac{\partial S_c(\bm{A})}{\partial A^k_{ij}}\quad 
\label{eqn::gradcam}
\end{equation}
where $w_k$ is the channel-wise weight that re-weights feature maps in different channels, and $g(\bm{A})$ denotes the explanatory function of the DNN output from $\bm{A}$. Detailed elaboration is provided in \textit{Appendix C}. Taking different layers as the attribution targets, we visualize the distribution of channel-wise average attribution gradients in Fig. \ref{fig::demo_avg}. It can be observed that the discrepancy of the weights $w_k$ is distinguishable in that OOD samples tend to produce more noisy and abnormal outliers compared to ID samples. Additionally, as the layers increase in depth, the magnitude of the average gradients also increases.

\begin{figure}[b]
  \centering
  \includegraphics[width=\textwidth]{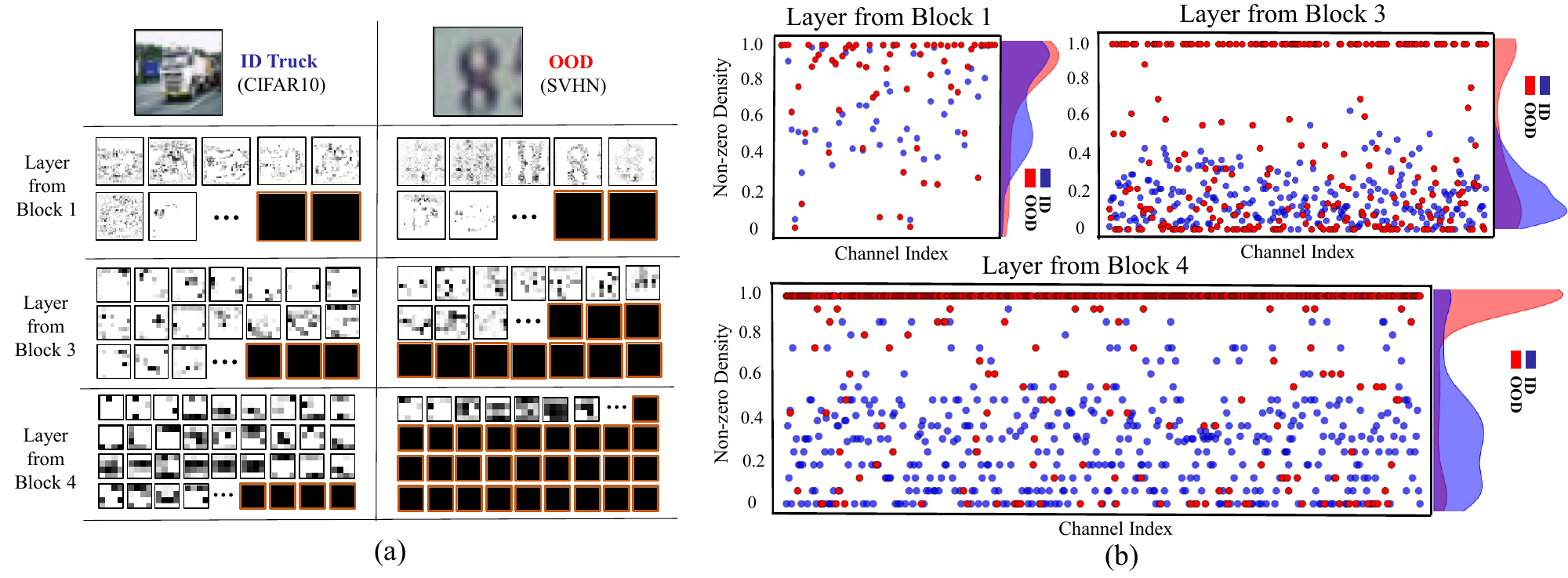}
  \caption{\textbf{Left (a):} Visualization of attribution gradients on feature maps. \textbf{Right (b):} Proportion of non-zero values across different channels. Each data point represents one single channel.}
  \label{fig::demo_zero}
\end{figure}

\textbf{Zero-deflation abnormality.} Then, we closely examine the abnormality that may arise in attribution gradients themselves due to distributional shifts. Fig. \ref{fig::demo_zero}(a) shows attribution gradients on feature maps across all channels at a specific layer. We observe  that the quantity of zero partial derivation $\frac{\partial S_c(\bm{A})}{\partial A^k_{ij}}$ in OOD is extremely less than ID, leading to a high occurrence of dense gradient matrices. As shown in Fig. \ref{fig::demo_zero}(b), this phenomenon is more pronounced in deeper layers,  indicating an abnormal behavior.

\subsection{Theoretical Explanation for Attribution Abnormality}

We consider a unified explanation for attribution algorithms with Taylor expansion. As proved in \cite{unified_attribution}, attribution algorithms are mathematically equivalent to the perspective that the network $c$-class output $S_c(\bm{z})$ is explained as a Taylor expansion model. For variables $\bm{z}=[z_1,...,z_n]$ (\eg., feature units to be attributed or inputs), 
here we perform $P$-order expansion of zero baseline output $S_c(\bm{0})$ at $\bm{z}$:
\begin{equation}
    S_c(\bm{0}) = S_c(\bm{z}) + \sum^{P}_{p=1} \sum_{i=1}^n  \frac{1}{p!} \frac{\partial^p S_c(\bm{z})}{\partial (z_i)^{p}} (z_i)^p +  \frac{1}{2!} \frac{\partial^2 S_c(\bm{z})}{\partial z_1 \partial z_2} z_1 z_2 + ... + R_P(\bm{z})
\end{equation}
where $R_P(\bm{z})$ denotes the remainder term for the $P$-order expansion. In our paper, we consider the feature values to be all zeros as the zero baseline, which is commonly adopted for analyzing gradient-based attribution algorithms. Then all terms can be represented by vector $\bm{\kappa}=[\kappa_1, ... , \kappa_n]\in \mathbb{N}^{n}$, where $\kappa_i \in \mathbb{N}$ reflects the integral degree of the input variable $z_i$ (\eg., $\kappa_i=1$ indicates the corresponding item only contains first-order partial derivative \wrt $z_i$). Thus we can represent the $c$-label output change caused by variables $\bm{z}$ as:
\begin{equation}
    | S_c(\bm{z}) - S_c(\bm{0}) | = | \sum_{p=1}^P \sum_{\bm{\kappa} \in \Omega, |\bm{\kappa}| = p} \mathcal{C}(\bm{\kappa}) \cdot \frac{\partial^{\kappa_1+\cdots+\kappa_n} S_c(\bm{z})}{\partial^{\kappa_1} z_1 \cdots \partial^{\kappa_n} z_n} (z_1)^{\kappa_1}\cdots(z_n)^{\kappa_n} + R_P(\bm{z}) |\\
\end{equation}
where $\mathcal{C(\bm{\kappa})}$ is a non-negative constant related to vector $\bm{\kappa}$. The expansion formula reflects the contribution of each variable $z_i$ to the $c$-label output change. Thus, we can attribute importance $\alpha_i$ to $z_i$ based on how much it contributes to such a change. Furthermore, the effect of $z_i$ to $S_c(\bm{z})$ can be decomposed into Taylor independent effect term $\phi_{i}(\bm{\kappa})$ and Taylor interaction effect term $\psi_{i}(\bm{\kappa})$. For independent term $\phi_{i}(\bm{\kappa})$, only $z_i$ is contained, where $\bm{\kappa} = [0,\cdots,\kappa_i,\cdots,0] \in \Omega_{\phi}$ and $\kappa_i>0$. And the overall effect of term $\psi_{i}(\bm{\kappa})$ is caused by the interactions between $z_i$ and other variables, where $\bm{\kappa}\in\Omega_{\psi}$ has at least two non-negative values and $\kappa_i>0$. Attribution methods are formulated as:
\begin{equation}
    \alpha_i = \sum_{p=1}^P \sum_{\bm{\kappa}\in\Omega_{\phi}, |\bm{\kappa}| = p} \omega_{i, \bm{\kappa}} \phi_i(\bm{\kappa}) + \sum_{p=1}^P \sum_{\bm{\kappa}\in\Omega_{\psi}, |\bm{\kappa}| = p} \omega_{i, \bm{\kappa}} \psi_i(\bm{\kappa})
\label{eqn::attn_reformulate}
\end{equation}
where $\omega_{i, \bm{\kappa}}$ denotes the ratio of a specific term (either the independent term or the interaction term) allocated to $\alpha_i$.

\textbf{Attribution abnormality in zero importance.} A reliable attribution result requires accurate identification of the features that are useful for the output. Here, we consider the Null-player axiom \cite{zero_attribution} (see \textit{Appendix E}), which states that in the reliable attribution, a feature should be considered as having zero importance when it makes no contribution to the model's output. In other words, \textit{if a feature does not contribute to the model's prediction, it should be considered as having zero importance.} 


\textbf{Proposition 1.} \textit{Given input variables $\bm{z}$, for one variable $z_i\in\bm{z}$ to be attributed, if $\frac{\partial S_c(\bm{z})}{\partial z_i}$ is zero throughout the analysis, then $\alpha_i=0$ always holds.}

Given variables $\bm{z}$ in one analysis, it is assumed that the partial derivative function \wrt $z_i$ is a constant zero. As shown in Eq. \ref{eqn:prop1},  all independent and interaction terms related to $z_i$ are zero. Thus, $z_i$ is of zero importance to the prediction. This is, zero attribution gradient values will directly impact the final attribution result.

\begin{equation}
    \frac{\partial S_c(\bm{z})}{\partial z_i} = 0 \Rightarrow  \frac{\partial^{\kappa_1+\cdots+\kappa_n} S_c(\bm{z})}{\partial^{\kappa_1} z_1 \cdots \partial^{\kappa_n} z_n}=0, \kappa_i>0 \Rightarrow \phi_i(\bm{\kappa})=\psi_i(\bm{\kappa})=0 \Rightarrow \alpha_i = 0
\label{eqn:prop1}
\end{equation}
This provides us with an explanatory perspective for our observation --- visual explanation for OOD data tends to be messy and unreliable due to the model's uncertainty about the unknown distribution, resulting in an abundance of intricate non-zero importance attributions.

\textbf{Attribution abnormality in gradient-based weights.} Following Eq. \ref{eqn::attn_reformulate}, GradCAM in Eq. \ref{eqn::gradcam} can be reformulated in form that includes only the first-order Taylor independent terms (see \textit{Appendix D} for the proof), where $\mathbf{\bm{\kappa}} = [0,...,\kappa_i=1,...0]$ is a one-hot vector, and $\kappa_j=0$ if $j \neq i$. This simplifies our analysis of the abnormality in weights, focusing solely on the correlation between first-order partial derivatives and the attribution result to reflect the uncertainty on each independent feature.

\section{GAIA: A Simple and Effective Framework for Abnormality Aggregation}

We propose our GAIA framework, which aggregates the channel-wise average abnormality (GAIA-A) or the zero-deflation abnormality (GAIA-Z) for out-of-distribution detection. 

\textbf{Abnormality aggregation from label space.} General attribution algorithms focus on the final predictive output $S_c(\bm{A})$, where $c=\text{argmax}_{c_{i} \in C} S_{c_{i}}(\bm{A})$. This is adequate for the zero-deflation abnormality as we aim to ascertain the model's confidence in interpreting its own classification result. While for the channel-wise average abnormality, our aspiration is to gather abnormalities from a broader label space. Hence, all outputs in the ID label space are informative for collecting the model's tendency towards identifying samples as ID categories. For GAIA-A, we fuse all the outputs with $\text{log}( \text{softmax}(\cdot))$:
\begin{equation}
    \frac{\partial \mathcal{S}(\bm{A})}{\partial A^{k}_{ij}} = \frac{\partial \sum_{c\in C} \text{log softmax}(S_c(\bm{A}))}{\partial A^k_{ij}}
    \label{eqn:fusion}
\end{equation}
This strategy first accumulates the model's outputs and simultaneously performs backpropagation \wrt the features $A^k_{ij}$. It is more efficient compared to individually backpropagating through each category and then accumulating them, which is impractical in scenarios with large label space (\eg., 1000 categories in ImageNet). Furthermore, we find that GAIA-A can be enhanced with a two-stage fusion strategy. Let us denote the neural network prediction function based on input feature variables $\bm{A}\in \mathbb{R}^{K \times W \times H}$ by $S_c(\bm{A})=\Psi(\bm{A}_{\text{last}}, \Theta_{\Psi})$, where $\Psi(\cdot)$ represents the classification function and $\bm{A}_{\text{last}}\in\mathbb{R}^{1\times W_{\text{last}} \times H_{\text{last}}}$ is the feature map at the last layer. Then the network feature extraction function is defined as $\Phi(\cdot)$ and $\bm{A}_{\text{last}} = \Phi(\bm{A}, \Theta_{\Phi})$. In our methods, we consider the gradient matrix on the $\bm{A}_{\text{last}}$ and the inner feature map $\bm{A}^{k}\in \mathbb{R}^{1\times W \times H}$ ($k$-th channel from $\bm{A}$) separately, with the former regarded as the output component $\nabla \bm{A}_{\text{last}}$ and the latter as the inner component $\nabla \bm{A}^k$:
\begin{equation}
\begin{aligned}
&\nabla \bm{A}^k = \frac{\partial \Phi(\bm{A}, \Theta_{\Phi})}{\partial \bm{A}^k}\\
&\nabla \bm{A}_{\text{last}} = \frac{\partial \mathcal{S}(\bm{A}_{\text{last}})}{\partial \bm{A}_{\text{last}}} = \frac{\partial \sum_{c\in C} \text{log softmax}(S_c(\bm{A}_{\text{last}}))}{\partial \bm{A}_{\text{last}}}
\end{aligned}
\label{eqn:twostage_fusion}
\end{equation}

\textbf{Abnormality aggregation from input space.} We start by defining the anormalies expectation on $k$-th channel feature map at $l$ layer as $\bm{A}^{kl} \in \mathbb{R}^{1\times W \times H}$. The zero-deflation abnormality can be described as the non-zero density of $\bm{A}^{kl}$:
\begin{equation}
    \mathbb{E}[\epsilon|\bm{A}^{kl}] = \frac{1}{W \times H} \mid \{A^{kl}_{ij} \quad|\quad \frac{\partial S_c(\bm{A}^{kl})}{\partial A^{kl}_{ij}}\neq0\} \mid
    \label{equ:zero_exp}
\end{equation}

For the channel-wise average abnormality, we observed that average gradients on $\bm{A}_{\text{last}}$ from the output component and the average attribution gradients obtained from the inner component exhibit opposite behaviors in terms of ID and OOD data \textit{(We discuss its effectiveness in Section \ref{sec::ablation} and provide theoretical analysis in \textit{Appendix F})}. Consequently, we use division to get the expectation of the final fusion channel-wise average abnormality abnormality:

\begin{equation}
    \mathbb{E}[\epsilon|\bm{A}^{kl}] = \frac{\mathbb{E}_{\text{inner}}[\epsilon|\bm{A}^{kl}]}{\sqrt{\mathbb{E}_{\text{output}}[\epsilon|\bm{A}_{\text{last}}]}} = \frac{ \mid \frac{1}{W \times H} \sum_{g^{kl} \in \nabla \bm{A}^{kl}}  g^{kl} \mid } { \mid \frac{1}{W_{\text{last}} \times H_{\text{last}}} \sum_{g_{\text{last}} \in \nabla \bm{A}_{\text{last}}} g_{\text{last}} \mid^{\frac{1}{2}}}
    \label{equ:avg_exp}
\end{equation}


Consider networks have $L$ layers to be utilized, and each layer has $K_l$ channels. Our framework accumulates them into an abnormality matrix $\bm{\Lambda} \in \mathbb{R}^{L \times K_m}$, where $K_m = \text{max} \{K_i | 1 \leq i \leq L\}$ and $\Lambda_{ij} = 0$ if $j > K_i$. Then, we use the Frobenius norm as a non-parameter measuring score to represent the global abnormality. For instance, assuming $K_m = K_L$, $\| \bm{\Lambda} \|_F$ is calculated as:
\begin{equation}
    \| \bm{\Lambda} \|_{F} = \begin{VNiceMatrix} \mathbb{E}[\epsilon|\bm{A}^{1,1}] & \cdots & \mathbb{E}[\epsilon|\bm{A}^{1,K_1}] & 0 & \cdots & 0\\ \Vdots && \Vdots   &&& \Vdots \\ \mathbb{E}[\epsilon|\bm{A}^{L,1}] & \cdots & \mathbb{E}[\epsilon|\bm{A}^{L,K_{1}}] & \Cdots & \Cdots & \mathbb{E}[\epsilon|\bm{A}^{L,K_{m}}]\end{VNiceMatrix}_{F} = \sqrt{\sum_i^L \sum_j^{K_m} (\mathbb{E}[\epsilon|\bm{A}^{i,j}])^2}
\label{equ:norm_score}
\end{equation} 
 
The overall process are formulized in Algorithm \ref{alg:GAIA}.
\begin{algorithm}[!htbp]
	\caption{GAIA}
	\label{alg:GAIA}
	\KwIn{Test sample $x$; Fixed model $f_{\theta}$.}
	\KwOut{OOD score $\Delta(x)$.}  
	\BlankLine
	
	Compute label output set $\{S_c(x) | c \in C\}$ by $f_{\theta}(x)$;
 
    Backpropagate attribution gradients by $\frac{\partial S_c(\bm{A}^{kl})}{\partial A^{kl}_{ij}}$ (GAIA-Z) or Eq. \ref{eqn:twostage_fusion} (GAIA-A);

    Calculate $\mathbb{E}[\epsilon|\bm{A}^{kl}]$ by Eq. \ref{equ:zero_exp} (GAIA-Z) or Eq. \ref{equ:avg_exp} (GAIA-A);

    Calculate global abnormality $\| \bm{\Lambda} \|_{F}$ by Eq. \ref{equ:norm_score};

    \Return $\| \bm{\Lambda} \|_{F}$ as OOD score $\Delta(x)$.
\end{algorithm}
\vspace{-0.5cm}
\section{Experiments}

In this section, we describe our experimental setup in Section \ref{sec::setup}. Then, we demonstrate the effectiveness of our method on the large-scale ImageNet-1K benchmark \cite{benchmark_imagenet}  and the CIFAR benchmarks \cite{MaxConfidenceScore} in Section \ref{sec::main_results}. We also conduct ablation studies in Section \ref{sec::ablation}.

\begin{table*}[t]
\centering
\resizebox{0.8\textwidth}{!}{
\begin{tabular}{c|l|cc|cc|cc|cc|cc|cc}
\toprule
\multirow{3}{*}{\textbf{\begin{tabular}[c]{@{}c@{}}ID\\ Datasets\end{tabular}}} & \multicolumn{1}{c|}{\multirow{3}{*}{\textbf{Methods}}} & \multicolumn{2}{c|}{\textbf{SVHN}}    & \multicolumn{2}{c|}{\textbf{TinyImageNet}}  & \multicolumn{2}{c|}{\textbf{LSUN}}     & \multicolumn{2}{c|}{\textbf{Places}} & \multicolumn{2}{c|}{\textbf{Textures}}    & \multicolumn{2}{c}{\textbf{Average}}    
\\ \cline{3-14} & \multicolumn{1}{c|}{}       & \multirow{2}{*}{\small{FPR95}}              & \multirow{2}{*}{\small{AUROC}}         & \multirow{2}{*}{\small{FPR95}}   & \multirow{2}{*}{\small{AUROC}}             & \multirow{2}{*}{\small{FPR95}}                & \multirow{2}{*}{\small{AUROC}}               & \multirow{2}{*}{\small{FPR95}}              & \multirow{2}{*}{\small{AUROC}}                 & \multirow{2}{*}{\small{FPR95}}              & \multirow{2}{*}{\small{AUROC}} &\multirow{2}{*}{\small{FPR95}}              & \multirow{2}{*}{\small{AUROC}}\\
& \multicolumn{1}{c|}{}                                 &  \small{($\downarrow$)} & (\small{$\uparrow$}) & (\small{$\downarrow$}) & (\small{$\uparrow$}) & (\small{$\downarrow$}) & (\small{$\uparrow$}) & (\small{$\downarrow$}) & (\small{$\uparrow$}) & (\small{$\downarrow$}) & (\small{$\uparrow$}) & (\small{$\downarrow$}) & (\small{$\uparrow$})\\ 
\midrule
                                
\multirow{11}{*}{\begin{tabular}[c]{@{}c@{}}CIFAR10\\ (ResNet34)\end{tabular}} 

& MSP \cite{MaxConfidenceScore}   & 61.03   & 89.01   & 53.11  & 85.79 & 46.79  & 90.63  & 43.71 & 91.88 & 48.28 & 90.08 & 50.58 & 89.48 \\
& ODIN \cite{ODIN}   & 50.74    &  92.09  & 39.82   & 92.62   & 33.34  & 94.17   & 36.53  & 93.18  & 45.00 & 91.11 & 41.09 & 92.63 \\

& Energy \cite{liu2020energy}   &  42.87  & 91.20     &  37.76  &  92.98    &  34.25   &  93.85 & 38.34  & 92.44  & 45.73 & 90.26 & 39.79 & 92.15 \\

& Mahalanobis \cite{ma_distance}     & 22.19      & 93.36       & 29.35     &  90.16  & 25.31     & 91.89       & 28.61    & 91.26  & 39.34 & 87.02 & 28.96 &  90.74 \\
& ReAct \cite{React}  &  24.60 & 92.39 & 33.68 & 89.71 & 19.15 & 93.78 & 23.69 & 92.78 & 32.61 & 89.27 & 26.75 & 91.59 \\

& GradNorm \cite{grad_norm}   & 62.47  & 76.08  & 73.00  & 65.21  & 59.38  & 72.97   & 58.93   & 75.36 & 67.77 & 67.41 & 64.31 & 71.41   \\

& KNN \cite{KNN}  &  32.03 & 95.28 & 29.56 & 95.44 & 27.42 & 95.92 & 41.77 & 93.26 & 35.41 & 94.87 & 33.24 & 94.95 \\

& Rankfeat \cite{rankfeat}  &  84.58 & 72.99 & 50.20 & 89.84 & 41.63 & 91.97 & 67.79 & 82.64 & 68.12 & 80.67 & 62.46 & 83.62 \\

& ASH-P@70 \cite{ASH}  &  23.11 & 95.53 & 29.78 & 93.71 & 22.72 & 95.33 & 25.27 & 94.35 & 30.92 & 93.08 & 26.36 & 94.40 \\

\cmidrule{2-14}
&\cellcolor{gray!25} \textbf{GAIA-Z (Ours)} &\cellcolor{gray!25} \textbf{2.47} &\cellcolor{gray!25} \textbf{99.49} &\cellcolor{gray!25} \textbf{6.26} &\cellcolor{gray!25} \textbf{98.63} &\cellcolor{gray!25} \textbf{2.48} &\cellcolor{gray!25} \textbf{99.43} &\cellcolor{gray!25} \textbf{2.27} &\cellcolor{gray!25} \textbf{99.50} &\cellcolor{gray!25} \textbf{2.84} &\cellcolor{gray!25} \textbf{99.36} &\cellcolor{gray!25} \textbf{3.26} &\cellcolor{gray!25} \textbf{99.28}\\
&\cellcolor{gray!25} \textbf{GAIA-A (Ours)} &\cellcolor{gray!25} \underline{14.44} &\cellcolor{gray!25} \underline{97.12} &\cellcolor{gray!25} \underline{16.45} &\cellcolor{gray!25} \underline{97.07} &\cellcolor{gray!25} \underline{9.10} &\cellcolor{gray!25} \underline{98.10} &\cellcolor{gray!25} \underline{11.06} &\cellcolor{gray!25} \underline{97.82} &\cellcolor{gray!25} \underline{12.62} &\cellcolor{gray!25} \underline{97.54} &\cellcolor{gray!25} \underline{12.73} &\cellcolor{gray!25} \underline{97.53}\\
\midrule
\multirow{11}{*}{\begin{tabular}[c]{@{}c@{}}CIFAR10\\ (WRN40)\end{tabular}} 

& MSP \cite{MaxConfidenceScore}   & 40.51   & 92.70   & 50.05  & 86.99 & 38.90  & 91.34  & 45.41 & 89.58 & 56.42 & 84.57 & 46.26 & 89.04 \\
& ODIN \cite{ODIN}   & \underline{16.11}    &  \underline{96.91}  & 44.18   & 89.66   & 33.37  & 93.45   & 40.30  & 91.31  & 51.51 & 87.71 & 37.09 & 91.81 \\
             
& Energy \cite{liu2020energy}   & 19.94    &  95.80  & 41.70   & 90.04   & 37.95  & 91.44   & 44.88  & 89.67  & 55.89 & 84.58 & 40.07 & 90.31 \\

& Mahalanobis \cite{ma_distance}     & 21.63      & 94.99       & 42.86     & 89.77   & 46.87     & 86.58       & 45.39    & 89.47  & 48.06 & 88.65 & 40.96 &  89.89 \\
& ReAct \cite{React}  &  20.05 & 95.87 & 41.32 & 90.29 & 37.81 & 91.57 & 44.28 & 89.77 & 54.88 & 85.54 & 39.67 & 90.61 \\

& GradNorm \cite{grad_norm}  &  49.60 & 80.45 & 82.23 & 59.60 & 78.17 & 63.55 & 81.70 & 59.72 & 82.93 & 58.05 & 74.93 & 64.27 \\ 

& KNN \cite{KNN}  & 27.52 & 95.55 & \underline{38.14} & \underline{93.44} & 38.95 & 94.30 & 45.79 & 90.65 & 50.37 & 90.77 & 40.16 & 92.94 \\

& Rankfeat \cite{rankfeat}  &  60.02 & 72.03 & 72.33 & 63.24 & 52.17 & 83.24 & 78.43 & 61.27 & 86.22 & 51.97 & 69.83 & 66.35 \\

& ASH-P@70 \cite{ASH}  &  19.94 & 95.80 & 41.70 & 90.04 & 37.96 & 91.44 & 44.53 & 89.75 & 55.69 & 84.71 & 39.97 & 90.35 \\

\cmidrule{2-14}
&\cellcolor{gray!25} \textbf{GAIA-Z (Ours)} &\cellcolor{gray!25} \textbf{4.05} &\cellcolor{gray!25} \textbf{99.17} &\cellcolor{gray!25} 53.31 &\cellcolor{gray!25} 90.59 &\cellcolor{gray!25} \textbf{12.40} &\cellcolor{gray!25} \textbf{97.92} &\cellcolor{gray!25} \textbf{7.76} &\cellcolor{gray!25} \textbf{98.59} &\cellcolor{gray!25} \textbf{12.30} &\cellcolor{gray!25} \textbf{97.34} &\cellcolor{gray!25} \textbf{17.96} &\cellcolor{gray!25} \textbf{96.72} \\
&\cellcolor{gray!25} \textbf{GAIA-A (Ours)} &\cellcolor{gray!25} 18.34 &\cellcolor{gray!25} 96.51 &\cellcolor{gray!25} \textbf{30.98} &\cellcolor{gray!25} \textbf{94.54} &\cellcolor{gray!25} \underline{12.73} &\cellcolor{gray!25} \underline{97.70} &\cellcolor{gray!25} \underline{16.94} &\cellcolor{gray!25} \underline{96.84} &\cellcolor{gray!25} \underline{14.93} &\cellcolor{gray!25} \underline{97.15} &\cellcolor{gray!25} \underline{18.78} &\cellcolor{gray!25} \underline{96.55} \\

\midrule

\multirow{11}{*}{\begin{tabular}[c]{@{}c@{}}CIFAR100\\ (ResNet34)\end{tabular}} 

& MSP \cite{MaxConfidenceScore}   & 86.21  & 74.13   & 75.21  & 79.31 & 83.58  & 72.80  & 87.19 & 70.60 & 82.00 & 74.46 & 82.84 & 74.26 \\
& ODIN \cite{ODIN}   & 89.34    &  70.21  & 70.00   & 81.44   & 83.80  & 71.37   & 88.10  & 67.69  & 81.81 & 72.66 & 82.61 & 72.67 \\

& Energy \cite{liu2020energy}   & 87.55    &  73.91  &  73.46   &  79.83   &  84.38  &  72.58   & 88.53  & 70.17  & 82.54 & 74.69 & 83.29 & 74.24 \\

& Mahalanobis \cite{ma_distance}     & 88.71      & 73.72       & 75.70     & 79.57   & 88.28     & 71.63       & 78.54    & 79.74  & 82.63 & 73.78 & 81.29 &  76.16 \\
& ReAct \cite{React}  & 77.53 & 83.17 & 71.18 & 78.60 & 73.36 & 84.37 & 78.41 & 80.12 & 72.06 & 82.54 & 74.51 & 81.76 \\

& GradNorm \cite{grad_norm}  &  90.70 & 65.95 & 80.12 & 61.44 & 82.62 & 58.10 & 92.29 & 64.35 & 85.89 & 52.48 & 86.32 & 60.46 \\ 

& KNN \cite{KNN}  &  73.34 & 80.06 & 69.24 & 82.17 & 76.98 & 78.36 & 86.76 & 71.53 & 79.95 & 69.24 & 77.25 & 76.27 \\

& Rankfeat \cite{rankfeat}  &  92.94 & 65.55 & 87.46 & 74.98 & 90.84 & 70.65 & 90.77 & 72.68 & 86.72 & 73.99 & 89.75 & 71.57 \\

& ASH-P@65 \cite{ASH}  &  81.21 & 79.46 & 74.26 & 81.17 & 82.84 & 74.93 & 85.49 & 72.91 & 79.70 & 77.33 & 80.70 & 77.16 \\

\cmidrule{2-14}
&\cellcolor{gray!25} \textbf{GAIA-Z (Ours)} &\cellcolor{gray!25} \textbf{15.73} &\cellcolor{gray!25} \textbf{97.06} &\cellcolor{gray!25} \textbf{63.85} &\cellcolor{gray!25} \textbf{89.17} &\cellcolor{gray!25} \textbf{33.33} &\cellcolor{gray!25} \textbf{94.18} &\cellcolor{gray!25} \textbf{16.78} &\cellcolor{gray!25} \textbf{97.17} &\cellcolor{gray!25} \textbf{15.82} &\cellcolor{gray!25} \textbf{97.09} &\cellcolor{gray!25} \textbf{29.10} &\cellcolor{gray!25} \textbf{94.93} \\
&\cellcolor{gray!25} \textbf{GAIA-A (Ours)} &\cellcolor{gray!25} \underline{68.02} &\cellcolor{gray!25} \underline{89.03} &\cellcolor{gray!25} \underline{68.61} &\cellcolor{gray!25} \underline{83.33} &\cellcolor{gray!25} \underline{71.24} &\cellcolor{gray!25} \underline{86.37} &\cellcolor{gray!25} \underline{73.15} &\cellcolor{gray!25} \underline{86.25} &\cellcolor{gray!25} \underline{63.81} &\cellcolor{gray!25} \underline{87.12} &\cellcolor{gray!25} \underline{68.97} &\cellcolor{gray!25} \underline{86.42} \\
\midrule

\multirow{11}{*}{\begin{tabular}[c]{@{}c@{}}CIFAR100\\ (WRN40)\end{tabular}} 

& MSP \cite{MaxConfidenceScore}   & 83.44  & 79.85   & 76.94  & 77.84 & 76.68  & 80.32  & 85.81 & 72.50 & 83.42 & 74.94 & 81.26 & 77.09 \\
& ODIN \cite{ODIN}   & 80.64    &  82.34  &  78.50   & 76.41   & 74.43  & 81.95   & 84.57  & 74.58  & 82.36 & 76.51 & 80.10 & 78.36 \\
& Energy \cite{liu2020energy}   & 84.58    &  79.72 & 76.77   & 77.90   & 76.32  & 80.45   & 86.13  & 72.35  & 83.95 & 74.83 & 81.55 & 77.05 \\
& Mahalanobis \cite{ma_distance}     & 82.36      & 81.07       & 82.95     & 79.20   & 74.76     & 81.16       & 82.44    & 76.06  & 83.72 & 76.93 & 80.97 &  78.34 \\
& ReAct \cite{React}  &  75.04 & 82.36 & 76.09 & 75.83 & 66.64 & 83.06 & 77.94 & 78.18 & 77.66 & 78.33 & 74.67 & 79.55 \\

& GradNorm \cite{grad_norm}  &  85.27 & 69.22 & 86.58 & 67.75 & 81.10 & 62.38 & 87.01 & 52.89 & 89.41 & 51.30 & 85.89 & 60.71 \\ 

& KNN \cite{KNN}  &  46.88 & 88.97 & \underline{70.88} & \underline{82.86} & 68.92 & 76.83 & 83.57 & 69.64 & 60.41 & 83.66 & 66.13 & 80.39 \\

& Rankfeat \cite{rankfeat}  &  80.39 & 77.10 & 94.58 & 52.35 & 91.63 & 61.89 & 86.83 & 67.71 & 88.00 & 67.36 & 88.29 & 65.28 \\

& ASH-P@70 \cite{ASH}  &  81.20 & 80.99 & 76.24 & 77.92 & 74.78 & 81.06 & 84.81 & 73.78 & 81.97 & 76.12 & 79.80 & 77.97 \\

\cmidrule{2-14}
&\cellcolor{gray!25} \textbf{GAIA-Z (Ours)} &\cellcolor{gray!25} \textbf{15.19} &\cellcolor{gray!25} \textbf{97.19} &\cellcolor{gray!25} {87.06} &\cellcolor{gray!25} {73.42} &\cellcolor{gray!25} \underline{37.97} &\cellcolor{gray!25} \underline{91.59} &\cellcolor{gray!25} \textbf{25.64} &\cellcolor{gray!25} \underline{95.26} &\cellcolor{gray!25} \textbf{27.29} &\cellcolor{gray!25} \underline{94.05} &\cellcolor{gray!25} \underline{38.63} &\cellcolor{gray!25} \underline{90.30} \\
&\cellcolor{gray!25} \textbf{GAIA-A (Ours)} &\cellcolor{gray!25} \underline{35.49} &\cellcolor{gray!25} \underline{93.60} &\cellcolor{gray!25} \textbf{53.37} &\cellcolor{gray!25} \textbf{89.86} &\cellcolor{gray!25} \textbf{33.52} &\cellcolor{gray!25} \textbf{93.86} &\cellcolor{gray!25} \underline{27.62} &\cellcolor{gray!25} \textbf{95.37} &\cellcolor{gray!25} \underline{31.44} &\cellcolor{gray!25} \textbf{94.16} &\cellcolor{gray!25} \textbf{36.29} &\cellcolor{gray!25} \textbf{93.37} \\

\bottomrule
\end{tabular}}
\caption{\textbf{Main Results on CIFAR Benchmarks \cite{MaxConfidenceScore}.} We evaluate on ResNet34 \cite{ResNet} and WideResnet40 \cite{wideResNet}, which are both pre-trained with cross-entropy loss. For Rankfeat and ASH, we choose their best performance. $\uparrow$ indicates larger values are better, while $\downarrow$ indicates smaller values are better. All values are percentages. The best results are in Bold and the second best results are underlined.}

\label{tab:main_cifar}
\end{table*}

\subsection{Setup}
\label{sec::setup}

\vspace{-0.1cm}
\textbf{Benchmarks.} In accordance with \cite{grad_norm, React, benchmark_imagenet, rankfeat}, we employ the large-scale ImageNet-1K benchmark \cite{benchmark_imagenet}, which offers a more realistic and challenging environment due to its use of high-resolution images and an large label space that encompasses 1,000 distinct categories. Four OOD datasets in this benchmark are from iNaturalist \cite{van2018inaturalist}, SUN \cite{xiao2010sun}, Places \cite{zhou2017places} and Textures \cite{cimpoi2014describing}, including fine-grained images, scene-oriented images, and textural images. We also evaluate CIFAR10 and CIFAR100 benchmarks \cite{MaxConfidenceScore}, which are routinely used in literature. Correspondingly, OOD datasets are SVHN \cite{SVHN}, TinyImageNet \cite{ODIN}, LSUN \cite{LSUN},  Places \cite{zhou2017places} and Textures \cite{cimpoi2014describing}.

\textbf{Baselines.} We consider various kinds of mainstream post-hoc OOD detection methods as baselines, including Maximum Softmax Probability (MSP) \cite{MaxConfidenceScore}, ODIN \cite{ODIN}, Energy-based method \cite{liu2020energy}, Mahalanobis \cite{ma_distance}, ReAct \cite{React}, GradNorm \cite{grad_norm}, Rankfeat \cite{rankfeat}, ASH \cite{ASH} and KNN \cite{KNN}. We use FPR95 (the false positive rate of OOD examples when the true positive rate of ID examples is 95\%) and AUROC (the area under the receiver operating characteristic curve) as evaluation metrics.

\vspace{-0.3cm}
\subsection{Main Results}
\label{sec::main_results}
\vspace{-0.2cm}

In our main results, all methods can be directly used for pre-trained models and for a fair comparison, auxiliary OOD data is unavailable for tuning.

\textbf{Evaluation on CIFAR benchmarks.} In Tab. \ref{tab:main_cifar}, we evaluate GAIA methods on CIFAR10 and CIFAR100 benchmarks. 
The results show that both GAIA-A and GAIA-Z exhibit superior performance. And we also note that advanced post-hoc methods such as Rankfeat and Gradnorm tend to encounter performance degradations on limited label space with small architectures. 
For ID dataset CIFAR10, baseline ASH performs the best with an average FPR95 of 26.36\% on ResNet34 and ODIN performs 37.09\% on WideResNet40 (WRN40). Our method GAIA-Z significantly outperforms ASH on ResNet34 by \textbf{23.10\% } improvement and outperforms ODIN on WideResNet by \textbf{19.13\%} improvement. Moreover, GAIA-A achieves the second best performance after GAIA-Z. For CIFAR100, GAIA-Z attains an average FPR95 of 29.10\% and average AUROC of 94.93\% on ResNet34, surpassing the best baseline ReAct by a margin of \textbf{45.41\%} FPR95 and \textbf{13.17\%} AUROC. GAIA-Z achieves surprising performance on CIFAR benchmarks by utilizing the zero-deflation abnormality.

\begin{table*}[!t]
\centering
\resizebox{0.9\textwidth}{!}{
\begin{tabular}{c|l|cc|cc|cc|cc|cc}
\toprule
\multirow{3}{*}{\textbf{\begin{tabular}[c]{@{}c@{}}Methods\\ Space\end{tabular}}} & \multicolumn{1}{c|}{\multirow{3}{*}{\textbf{Methods}}} & \multicolumn{2}{c|}{\textbf{iNaturalist}}    & \multicolumn{2}{c|}{\textbf{SUN}}  & \multicolumn{2}{c|}{\textbf{Places}}    & \multicolumn{2}{c|}{\textbf{Textures}}       & \multicolumn{2}{c}{\textbf{Average}}    
\\ \cline{3-12} & \multicolumn{1}{c|}{}       & \multirow{2}{*}{FPR95 $\downarrow$}              & \multirow{2}{*}{AUROC $\uparrow$}         & \multirow{2}{*}{FPR95 $\downarrow$}   & \multirow{2}{*}{AUROC $\uparrow$}             & \multirow{2}{*}{FPR95 $\downarrow$}                & \multirow{2}{*}{AUROC $\uparrow$}               & \multirow{2}{*}{FPR95 $\downarrow$}              & \multirow{2}{*}{AUROC $\uparrow$}                 & \multirow{2}{*}{FPR95 $\downarrow$}              & \multirow{2}{*}{AUROC $\uparrow$}     \\
& \multicolumn{1}{c|}{}                                 &  &  &  &  &  &  &  &  &  &   \\ \midrule
                                
\multirow{3}{*}{Output} 

& MSP \cite{MaxConfidenceScore}   & 63.69   & 87.59   & 79.98  & 78.34 & 81.44  & 76.76  & 82.73  & 74.45 & 76.96   & 79.29                \\
& ODIN \cite{ODIN}   & \textbf{62.69}    & \textbf{89.36}    & 71.67   & 83.92   & 76.27  & 80.67   & 81.31  & \textbf{76.30}   & 72.99  & 82.56                \\
& Energy \cite{liu2020energy}       & 64.91   & 88.48     & \textbf{65.33}   & \textbf{85.32}    & \textbf{73.02}   & \textbf{81.37}  & \textbf{80.87}  & 75.79     & \textbf{71.03}   & \textbf{82.74}    \\
  \midrule
\multirow{6}{*}{Feature} 
& Mahalanobis \cite{ma_distance}     & 96.34      & 46.33       & 88.43     & 65.20   & 89.75     & 64.46       & 52.23    & 72.10                 & 81.69                & 62.02      \\
& ReAct \cite{React}  &  44.52 & 91.81 & 52.71 & 90.16 & 62.66 & 87.83 & 70.73 & 76.85 & 57.66 & 86.67 \\

& KNN \cite{KNN}  &  59.08 & 86.20 & 69.53 & 80.10 & 77.09 & 74.87 &  \textbf{11.56} & \textbf{97.18} & 54.32 & 84.59 \\

& Rankfeat (Block4)\cite{rankfeat}  &  46.54 & 81.49 & \textbf{27.88} & 92.18 & \textbf{38.26} & 88.34 & 46.06 & 89.33 & 39.69 & 87.84 \\

& Rankfeat (Block3+4)\cite{rankfeat}  &  41.31 & 91.91 & 29.27 &\textbf{94.07} & 39.34 & \textbf{90.93} &  37.29 & 91.70 & 36.80 & 92.15 \\

& ASH-B@90 \cite{ASH}  &  \textbf{22.22} & \textbf{96.15} & 35.43 & 92.53 & 47.73 & 89.61 &  23.33 & 95.43 & \textbf{32.18} & \textbf{93.43} \\

\midrule
\multirow{3}{*}{Gradient}                    
& GradNorm \cite{grad_norm}   & 50.03  & 90.33  & 46.48  & 89.03  & 60.86  & 84.82     & 61.42   & 81.07   & 54.70  & 86.31  \\
& \cellcolor{gray!25}\textbf{GAIA-A (Ours)} & \cellcolor{gray!25}\textbf{29.47} & \cellcolor{gray!25}\textbf{93.52} & \cellcolor{gray!25}\textbf{31.24} & \cellcolor{gray!25}\textbf{92.42} & \cellcolor{gray!25}\textbf{48.55} & \cellcolor{gray!25}\textbf{88.94} & \cellcolor{gray!25}40.41 & \cellcolor{gray!25}92.71 & \cellcolor{gray!25}\textbf{37.42} & \cellcolor{gray!25}\textbf{91.90} \\
& \cellcolor{gray!25}\textbf{GAIA-Z (Ours)} & \cellcolor{gray!25}65.09 & \cellcolor{gray!25}84.15 & \cellcolor{gray!25}64.23 & \cellcolor{gray!25}84.31 & \cellcolor{gray!25}71.02 & \cellcolor{gray!25}81.16 & \cellcolor{gray!25}\textbf{11.32} & \cellcolor{gray!25}\textbf{97.93} & \cellcolor{gray!25}52.92 &  \cellcolor{gray!25}86.89\\
  
\bottomrule
\end{tabular}
}
\caption{\textbf{Main Results on ImageNet-1K \cite{benchmark_imagenet}.} OOD detection performance comparison between GAIA and advanced baselines on pre-trained Google BiT-S \cite{resnetv2} model.  Our methods only use layers from Block4 and all methods are post hoc that can be directly used for pre-trained models. The best results for each Methods Space are all in Bold.}

\label{table:main_imagenet}
\end{table*}

\textbf{Evaluation on ImageNet-1K benchmark.} In Tab. \ref{table:main_imagenet}, we compare GAIA with other post hoc baselines on pre-trained Google BiT-S model \cite{resnetv2}. For our methods, both GAIA-A and GAIA-Z use layers from the last block (\textbf{Block4}), and no hyperparameters are required. GAIA-A performs well with an average FPR95 of 37.42\% and an average AUROC of 91.90\%. Compared to other gradient-based OOD methods, GAIA-A outperforms GradNorm by \textbf{17.28\%} in FPR95. Besides, GAIA-Z excels in handling the OOD dataset of textures with $11.32\%$ FPR95, despite not achieving the best overall performance. While ASH achieves competitive results on the ImageNet dataset through careful parameter tuning, it is highly sensitive to its hyperparameters and lacks empirical parameters. In contrast, GAIA methods don't require parameter adjustments and directly achieve good results.

\subsection{Ablation Studies}
\label{sec::ablation}

Our ablation study begins by validating the effectiveness of each step of the methods. We first verify the effect of the Frobenius norm (2-norm). Then we explore the aggregation's effectiveness on the label space and the input space.

\begin{figure}[!hbpt]
\vspace{-0.2cm}
  \centering
  \includegraphics[width=0.9\textwidth]{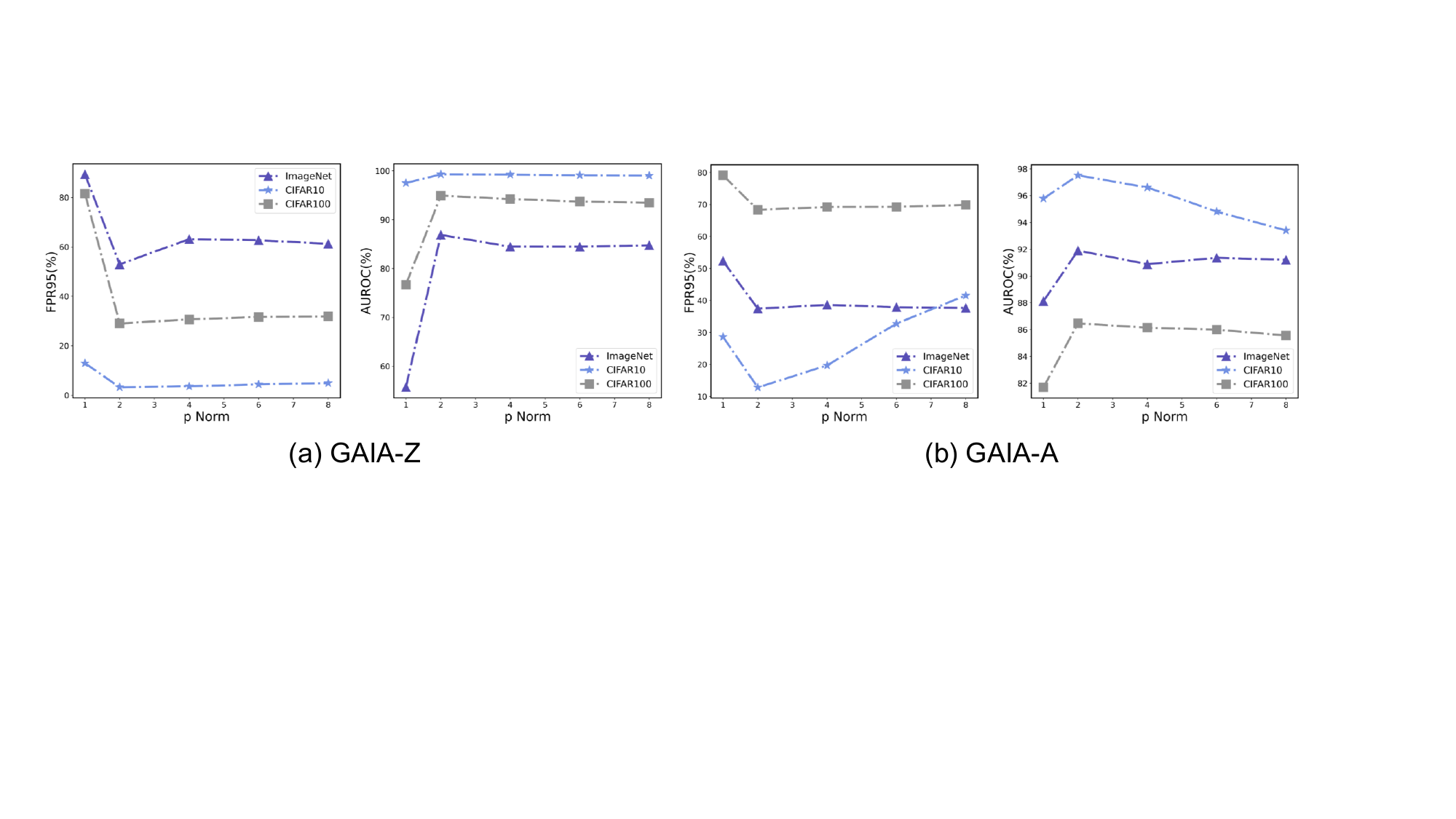}
  \vspace{-0.2cm}
  \caption{Ablation studies on Frobenius norm of matrix $\bm{\Lambda$}.}
  \label{fig::exp_norm}
\end{figure}

\textbf{Influence of Frobenius norm.} In Eq. \ref{equ:norm_score}, we use $\|\bm{\Lambda}\|_{F}$ to calculate the final OOD score. To verify its effectiveness, we evaluate different norms of $\bm{\Lambda}$ on the above three benchmarks. As shown in Fig. \ref{fig::exp_norm}, the Frobenius norm performs the best. Compared to 1-norm, Frobenius norm particularly demonstrates significant improvements. This is because the Frobenius norm can exclude the influence of numerous smaller values. As the number of layers in the model increases, the accumulation of insignificant small values in the shallow layers can weaken the scoring impact of extreme values OOD data. However, we can observe that as the value of $p$ increases, the influence of extreme values will also be affected.

\begin{figure}[t]
  \centering
  \includegraphics[width=0.85\textwidth]{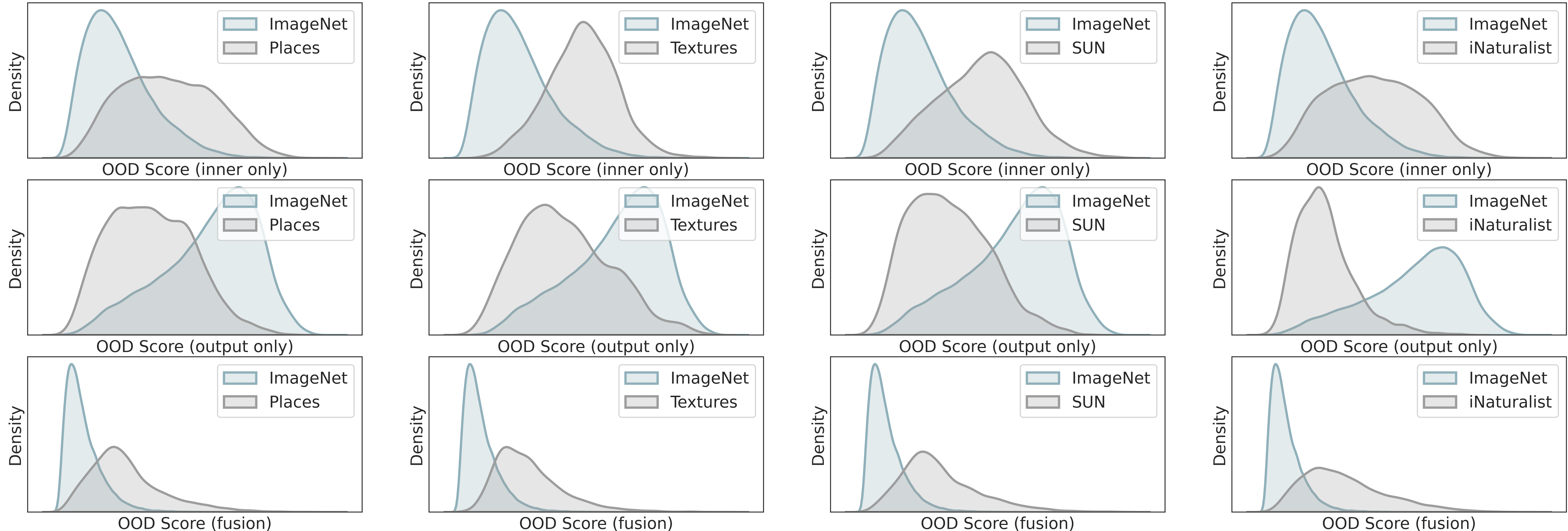}
  \caption{The distribution of the OOD scores in three settings (\textit{inner only}, \textit{output only} and \textit{fusion}). All scores are non-negative for comparison.}
  \label{fig::all_fusion}
\end{figure}

\textbf{Influence of label space aggregation.} In GAIA-A, we employ division to fuse the inner component $\mathbb{E}_{\text{inner}}[\epsilon | \bm{A}^{kl}]$ and the output component $\mathbb{E}_{\text{output}}[\epsilon|\bm{A}^{kl}]$ to obtain the final OOD scores. As shown in Fig. \ref{fig::all_fusion}, we visualize the score distributions of the individual components and the fused scores, and observe that the performance of the inner and output components in OOD and ID data are contrasting. After dividing and merging the two components, the fusion resulted in a greater concentration of ID data, tending towards a narrower distribution. However, the impact on the distribution of OOD data was relatively minor, thereby widening the score differences between them. In Tab. \ref{table:ablation_fusion}, we compare the OOD detection performance with and without (w/o) the fusion strategy. Experiments demonstrated a improvement with the implementation of this strategy. 

\begin{table*}[!hp]
\centering
\resizebox{0.9\textwidth}{!}{
\begin{tabular}{l|cc|cc|cc|cc|cc}
\toprule
\multicolumn{1}{c|}{\multirow{3}{*}{\textbf{Methods}}} & \multicolumn{2}{c|}{\textbf{iNaturalist}}    & \multicolumn{2}{c|}{\textbf{SUN}}  & \multicolumn{2}{c|}{\textbf{Places}}    & \multicolumn{2}{c|}{\textbf{Textures}}       & \multicolumn{2}{c}{\textbf{Average}}    
\\ \cline{2-11}       & \multirow{2}{*}{FPR95 $\downarrow$}              & \multirow{2}{*}{AUROC $\uparrow$}         & \multirow{2}{*}{FPR95 $\downarrow$}   & \multirow{2}{*}{AUROC $\uparrow$}             & \multirow{2}{*}{FPR95 $\downarrow$}                & \multirow{2}{*}{AUROC $\uparrow$}               & \multirow{2}{*}{FPR95 $\downarrow$}              & \multirow{2}{*}{AUROC $\uparrow$}                 & \multirow{2}{*}{FPR95 $\downarrow$}              & \multirow{2}{*}{AUROC $\uparrow$}     \\
& \multicolumn{1}{c}{}                                  &  &  &  &  &  &  &  &  &   \\ \midrule
w/o fusion (top 1 label) & 74.44 & 74.04 & 77.30 & 77.60 & 82.07 & 71.11 & 50.14 & 89.57 & 70.99 & 78.08\\
w/o fusion (output only) & 47.50 & 92.54 & 69.87 & 84.47 & 74.52 & 82.00 & 76.17 & 78.83 & 67.02 & 84.46 \\
w/o fusion (inner only) & 59.45 & 81.98 & 52.24 & 86.51 & 62.74 & 80.20 & 52.45 & 89.77 & 56.72 & 84.62 \\
\cellcolor{gray!25} \textbf{fusion (logsoftmax + division)} & \cellcolor{gray!25}\textbf{29.47} & \cellcolor{gray!25}\textbf{93.52} & \cellcolor{gray!25}\textbf{31.24} & \cellcolor{gray!25}\textbf{92.42} & \cellcolor{gray!25}\textbf{48.55} & \cellcolor{gray!25}\textbf{88.94} & \cellcolor{gray!25}\textbf{40.41} & \cellcolor{gray!25}\textbf{92.71} & \cellcolor{gray!25}\textbf{37.42} & \cellcolor{gray!25}\textbf{91.90} \\
  
\bottomrule
\end{tabular}
}
\caption{Ablation studies on fusion strategy. \textit{top 1 label} means utilizing the predictive output only.}
\label{table:ablation_fusion}
\end{table*}

\vspace{-0.1cm}
\textbf{Influence of input space aggregation across different layers (blocks).} Given that both ResNet34 and Google BiT-S models have four blocks, we analyze the performance of our methods across different blocks to elucidate the influence of feature layers. As shown in Tab. \ref{table:ablation_blocks}, deeper layers possess a higher power in distinguishing between ID and OOD data. It indicates that as the network becomes shallower, the feature maps progressively contain a diminishing amount of relevant information \wrt the prediction decision \cite{guidotti2018survey}. For CIFAR benchmarks, information from Block3+4 is sufficient for detection, and for ImageNet-1K benchmark, only using Block4 can achieve the best performance.

\begin{table*}[hp]
\centering
\resizebox{0.9\textwidth}{!}{
\begin{tabular}{c | cccc | cccc | cccc}
\toprule
\multicolumn{1}{c|}{\multirow{3}{*}{\textbf{Blocks}}} & \multicolumn{4}{c|}{\textbf{CIFAR10}}    & \multicolumn{4}{c|}{\textbf{CIFAR100}}  & \multicolumn{4}{c}{\textbf{ImageNet}} \\
&  \multicolumn{2}{c}{GAIA-A}              & \multicolumn{2}{c|}{GAIA-Z}        & \multicolumn{2}{c}{GAIA-A}   & \multicolumn{2}{c|}{GAIA-Z}            & \multicolumn{2}{c}{GAIA-A}                & \multicolumn{2}{c}{GAIA-Z} \\
& \small{FPR95$\downarrow$} & \small{AUROC$\uparrow$} & \small{FPR95$\downarrow$} & \small{AUROC$\uparrow$} & \small{FPR95$\downarrow$} & \small{AUROC$\uparrow$} & \small{FPR95$\downarrow$} & \small{AUROC$\uparrow$} & \small{FPR95$\downarrow$} & \small{AUROC$\uparrow$} & \small{FPR95$\downarrow$} & \small{AUROC$\uparrow$}\\
\midrule
Block 1 & 64.52 & 83.28 & 57.15 & 67.06 & 77.89 & 79.72 & 58.38 & 66.46 & 86.59 & 61.26 & 92.38 & 49.62\\
Block 2 & 62.19 & 86.26 & 50.71 & 86.69 & 77.30 & 80.50 & 52.40 & 86.96 & 87.21 & 58.39 & 88.56 & 58.64\\
Block 3 & 44.56 & 91.07 & 22.17 & 95.71 & 71.28 & 84.67 & 44.18 & 89.01 & 63.34 & 80.81 & 73.28 & 78.96\\
Block 4 & 12.90 & 97.54 & 6.42 & 98.75 & 69.16 & 86.40 & 49.97 & 91.28 & \textbf{37.42} & \textbf{91.90}  & \textbf{52.92} & \textbf{86.89}\\
Block 3+4 & \textbf{12.70} & \textbf{97.53} & 3.55 & 99.26 & \textbf{68.98} & \textbf{86.42} & \textbf{27.86} & \textbf{95.24} & 41.91 & 91.03 & 58.39 & 86.91\\
All blocks & 12.73 & 97.53 & \textbf{3.26} & \textbf{99.28} & 68.98 & 86.42 & 29.05 & 94.92 & 42.38 & 90.86 & 63.28 & 86.02\\

\bottomrule
\end{tabular}}
\caption{Ablation studies of the influence on different blocks with average FPR95 and AUROC.}
\label{table:ablation_blocks}
\end{table*}
\vspace{-0.3cm}
\section{Related Work}
\vspace{-0.1cm}
Among all attempts so far, post-hoc methods \cite{ma_distance, ODIN,liu2020energy, grad_norm, React, rankfeat} are preferable in the wild due to their advantages of being easy to use without modifying the training procedure and objective. 
An initial solution proposed by \citeauthor{MaxConfidenceScore}~\citep{MaxConfidenceScore} utilizes maximum softmax probability (MSP). While due to the tendency of networks to display overconfident softmax scores when predicting OOD inputs \cite{networks_fooled, OOD_overconfidence_hein2019relu}, it renders a non-trivial dilemma to separate ID and OOD data. Then ODIN \cite{ODIN} introduces temperature factors and input perturbations to enhance detection performance. In a different approach, Energy \cite{liu2020energy} is proposed to utilize the energy score as an informative indicator. ReAct \cite{React} proposes that OOD examples result in abnormal model activation and suggests clamping the activation values above a threshold. Rankfeat \cite{rankfeat} leverages the differences in singular value distributions, which still focuses on abnormal activations of the model. Another relevant study to this paper is gradient-based OOD detection. In the early work, ODIN \cite{ODIN} first implicitly utilizes gradients as perturbations to increase the softmax score of any given input. Recently, \citeauthor{lee2020gradients}~\citep{lee2020gradients}, \citeauthor{grad_norm}~\citep{grad_norm} and \citeauthor{gradient_igoe2022useful}~\citep{gradient_igoe2022useful} use the gradients of parameters as the measurement, which emphasizes the importance of the loss function. In this paper, we delve into investigating attribution abnormality and utilize attribution gradients for OOD detection.

\section{Discussion}

In this section, we discuss the comparison of our methods with other gradient-based OOD detection methods, as well as the limitation on transformer-based models.

\vspace{-0.2cm}
\subsection{Comparison with Other Gradient-based Methods} 

A crucial distinction between other gradient-based OOD detection methods and ours lies in the utilization of attribution methods to interpret the anomalous behavior of OOD examples. Specifically, we investigate and aggregate the abnormal patterns exhibited by attribution gradients at the feature level. Compared to ODIN \cite{ODIN}, GAIA directly leverages the uncertainty derived from the gradients of input features, providing a more intuitive and efficient solution. Furthermore, rather than focusing solely on the softmax output, we delve into the intermediate statistics to uncover more fundamental discrepancies. Compared to GradNorm \cite{grad_norm}, ExGrad \cite{gradient_igoe2022useful} and \citeauthor{lee2020gradients}~\cite{lee2020gradients}, our approaches focus on attribution gradients and demonstrate superior performance.  
The comparative performance is presented in Tab. \ref{table:ablation_gradient_methods1}. Additionally, GAIA supports batch processing, as the attribution gradients are independent for each input feature, while gradients of parameters are unique to the network. This means that our method can handle multiple samples simultaneously, providing a parallel processing advantage over these methods that can only process one sample at a time.

\begin{table*}[!hp]
\centering
\resizebox{0.8\textwidth}{!}{
\begin{tabular}{l|c|c|c|c|c|c}
\toprule
\multicolumn{1}{c|}{\multirow{2}{*}{\textbf{Methods}}} & \multicolumn{1}{c|}{\multirow{2}{*}{\textbf{Batch processing}}} &\multicolumn{1}{c|}{\textbf{iNaturalist}}    & \multicolumn{1}{c|}{\textbf{SUN}}  & \multicolumn{1}{c|}{\textbf{Places}}    & \multicolumn{1}{c|}{\textbf{Textures}}       & \multicolumn{1}{c}{\textbf{Average}}    
\\ \cline{3-7}                  &&   \small{AUROC $\uparrow$}          & \small{AUROC $\uparrow$}                   & \small{AUROC $\uparrow$}                     & \small{AUROC $\uparrow$}                        & \multirow{1}{*}\small{AUROC $\uparrow$}    \\  \midrule
Lee and AlRegib \cite{lee2020gradients} &&   72.30 &  82.61  & 74.00  & 84.16  & 78.27\\
GradNorm \cite{grad_norm} &&  90.33  & 89.03  & 84.82  & 81.07  & 86.31\\
ExGrad \cite{gradient_igoe2022useful} &&   76.90  & 66.60  & 68.90  & 65.10  & 69.40\\
\textbf{GAIA-A (Ours)} & $\checkmark$  & \textbf{93.52}  & \textbf{92.42}  & \textbf{88.94}  & 92.71  & \textbf{91.90}\\
\textbf{GAIA-Z (Ours)} & $\checkmark$  & 84.15  & 84.31  & 81.16  & \textbf{97.93} &  86.89\\
\bottomrule
\end{tabular}}
\caption{Comparison with other gradient-based methods. To ensure a fair comparison with Lee and AlRegib \cite{lee2020gradients}, the gradients of uniform noise are used as a surrogate, as suggested in \cite{grad_norm}.}
\label{table:ablation_gradient_methods1}
\end{table*}

\vspace{-0.2cm}
\subsection{Limitation on Transformer-based Models}

Newer models like Vision Transformers (ViT) \cite{vit}, which are based on transformers, excel in feature extraction. However, they may not align well with image-specific characteristics. For instance, ViTs employ positional encoding to capture spatial information, posing challenges for attribution. Due to this reason, existing attribution algorithms are rarely applied to ViTs, resulting in poorer performance for GAIA. 
While the attention mechanism in transformer-based models can also offer directions for visual explanations. In our future work, we will research the uncertainty in the attention matrix to enhance OOD detection performance on transformer-based models.

\vspace{-0.3cm}
\section{Conclusion}
\label{related_work}

This paper targets bridging the gap between OOD detection and visual interpretation by utilizing the uncertainty of a model in explaining its own predictions. We further examine how attribution gradients contribute to uncertain explanation outcomes and introduce two forms of abnormalities for OOD detection.  Then, we propose GAIA, a simple and effective framework for abnormality aggregation. The effectiveness of our framework is validated through experiments.

\textbf{Societal impact and limitations.} Through this work, we aim to provide a new perspective to improve the performance of OOD detection and ensure the safety and reliability of machine learning applications. However, the utilization of attribution gradients in this paper is relatively simplistic. We believe there is still significant research potential in this area. Moreover, the limitation on transformer-based models remains a topic for further investigation.

\section{Acknowledgement}

Research is supported by the Key Research and Development Program of Guangdong Province (grant No. 2021B0101400003). This work was done while Jinggang Chen was interning at Ping
An Technology  and the corresponding authors are Xiaoyang Qu and Jianzong Wang from Ping An Technology (Shenzhen) Co., Ltd.


\appendix

\bibliographystyle{unsrtnat}
\bibliography{neurips_2023}

\begin{thebibliography}{42}
\providecommand{\natexlab}[1]{#1}
\providecommand{\url}[1]{\texttt{#1}}
\expandafter\ifx\csname urlstyle\endcsname\relax
  \providecommand{\doi}[1]{doi: #1}\else
  \providecommand{\doi}{doi: \begingroup \urlstyle{rm}\Url}\fi

\bibitem[Huang et~al.(2020)Huang, Kroening, Ruan, Sharp, Sun, Thamo, Wu, and Yi]{huang2020survey}
Xiaowei Huang, Daniel Kroening, Wenjie Ruan, James Sharp, Youcheng Sun, Emese Thamo, Min Wu, and Xinping Yi.
\newblock A survey of safety and trustworthiness of deep neural networks: Verification, testing, adversarial attack and defence, and interpretability.
\newblock \emph{Computer Science Review}, 37:\penalty0 100270, 2020.

\bibitem[Litjens et~al.(2017)Litjens, Kooi, Bejnordi, Setio, Ciompi, Ghafoorian, Van Der~Laak, Van~Ginneken, and S{\'a}nchez]{medical}
Geert Litjens, Thijs Kooi, Babak~Ehteshami Bejnordi, Arnaud Arindra~Adiyoso Setio, Francesco Ciompi, Mohsen Ghafoorian, Jeroen~Awm Van Der~Laak, Bram Van~Ginneken, and Clara~I S{\'a}nchez.
\newblock A survey on deep learning in medical image analysis.
\newblock \emph{Medical Image Analysis}, 42:\penalty0 60--88, 2017.

\bibitem[Ozbayoglu et~al.(2020)Ozbayoglu, Gudelek, and Sezer]{fiance}
Ahmet~Murat Ozbayoglu, Mehmet~Ugur Gudelek, and Omer~Berat Sezer.
\newblock Deep learning for financial applications: A survey.
\newblock \emph{Applied Soft Computing}, 93:\penalty0 106384, 2020.

\bibitem[Hsu et~al.(2020)Hsu, Shen, Jin, and Kira]{baseline_G-ODIN}
Yen-Chang Hsu, Yilin Shen, Hongxia Jin, and Zsolt Kira.
\newblock Generalized odin: Detecting out-of-distribution image without learning from out-of-distribution data.
\newblock In \emph{Proceedings of the IEEE/CVF Conference on Computer Vision and Pattern Recognition}, pages 10951--10960, 2020.

\bibitem[Liu et~al.(2020)Liu, Wang, Owens, and Li]{liu2020energy}
Weitang Liu, Xiaoyun Wang, John Owens, and Yixuan Li.
\newblock Energy-based out-of-distribution detection.
\newblock \emph{Advances in Neural Information Processing Systems}, 33, 2020.

\bibitem[Zaeemzadeh et~al.(2021)Zaeemzadeh, Bisagno, Sambugaro, Conci, Rahnavard, and Shah]{1-D}
Alireza Zaeemzadeh, Niccolo Bisagno, Zeno Sambugaro, Nicola Conci, Nazanin Rahnavard, and Mubarak Shah.
\newblock Out-of-distribution detection using union of 1-dimensional subspaces.
\newblock In \emph{Proceedings of the IEEE/CVF Conference on Computer Vision and Pattern Recognition}, pages 9452--9461, 2021.

\bibitem[Hendrycks and Gimpel(2017)]{MaxConfidenceScore}
Dan Hendrycks and Kevin Gimpel.
\newblock A baseline for detecting misclassified and out-of-distribution examples in neural networks.
\newblock In \emph{International Conference on Learning Representations}, 2017.

\bibitem[Lakshminarayanan et~al.(2017)Lakshminarayanan, Pritzel, and Blundell]{lakshminarayanan2017simple}
Balaji Lakshminarayanan, Alexander Pritzel, and Charles Blundell.
\newblock Simple and scalable predictive uncertainty estimation using deep ensembles.
\newblock \emph{Advances in Neural Information Processing Systems}, 30, 2017.

\bibitem[Liang et~al.(2018)Liang, Li, and Srikant]{ODIN}
Shiyu Liang, Yixuan Li, and R.~Srikant.
\newblock Enhancing the reliability of out-of-distribution image detection in neural networks.
\newblock In \emph{International Conference on Learning Representations}, 2018.

\bibitem[Huang et~al.(2021)Huang, Geng, and Li]{grad_norm}
Rui Huang, Andrew Geng, and Yixuan Li.
\newblock On the importance of gradients for detecting distributional shifts in the wild.
\newblock \emph{Advances in Neural Information Processing Systems}, 34, 2021.

\bibitem[Sun et~al.(2021)Sun, Guo, and Li]{React}
Yiyou Sun, Chuan Guo, and Yixuan Li.
\newblock React: Out-of-distribution detection with rectified activations.
\newblock \emph{Advances in Neural Information Processing Systems}, 34, 2021.

\bibitem[Hendrycks et~al.(2018)Hendrycks, Mazeika, and Dietterich]{OE}
Dan Hendrycks, Mantas Mazeika, and Thomas Dietterich.
\newblock Deep anomaly detection with outlier exposure.
\newblock \emph{International Conference on Learning Representations}, 2018.

\bibitem[Sastry and Oore(2020)]{Gram}
Chandramouli~Shama Sastry and Sageev Oore.
\newblock Detecting out-of-distribution examples with gram matrices.
\newblock In \emph{International Conference on Machine Learning}, pages 8491--8501, 2020.

\bibitem[Lee et~al.(2018)Lee, Lee, Lee, and Shin]{ma_distance}
Kimin Lee, Kibok Lee, Honglak Lee, and Jinwoo Shin.
\newblock A simple unified framework for detecting out-of-distribution samples and adversarial attacks.
\newblock \emph{Advances in Neural Information Processing Systems}, 31, 2018.

\bibitem[Song et~al.(2022)Song, Sebe, and Wang]{rankfeat}
Yue Song, Nicu Sebe, and Wei Wang.
\newblock Rankfeat: Rank-1 feature removal for out-of-distribution detection.
\newblock \emph{Advances in Neural Information Processing Systems}, 2022.

\bibitem[Lee and AlRegib(2020)]{lee2020gradients}
Jinsol Lee and Ghassan AlRegib.
\newblock Gradients as a measure of uncertainty in neural networks.
\newblock In \emph{2020 IEEE International Conference on Image Processing (ICIP)}, pages 2416--2420, 2020.

\bibitem[Igoe et~al.(2022)Igoe, Chung, Char, and Schneider]{gradient_igoe2022useful}
Conor Igoe, Youngseog Chung, Ian Char, and Jeff Schneider.
\newblock How useful are gradients for ood detection really?
\newblock \emph{arXiv preprint arXiv:2205.10439}, 2022.

\bibitem[Simonyan et~al.(2013)Simonyan, Vedaldi, and Zisserman]{attribution_sensitivity}
Karen Simonyan, Andrea Vedaldi, and Andrew Zisserman.
\newblock Deep inside convolutional networks: Visualising image classification models and saliency maps.
\newblock \emph{International Conference on Learning Representations}, 2013.

\bibitem[Selvaraju et~al.(2017)Selvaraju, Cogswell, Das, Vedantam, Parikh, and Batra]{channel_mean}
Ramprasaath~R Selvaraju, Michael Cogswell, Abhishek Das, Ramakrishna Vedantam, Devi Parikh, and Dhruv Batra.
\newblock Grad-cam: Visual explanations from deep networks via gradient-based localization.
\newblock In \emph{Proceedings of the IEEE International Conference on Computer Vision}, pages 618--626, 2017.

\bibitem[Chattopadhay et~al.(2018)Chattopadhay, Sarkar, Howlader, and Balasubramanian]{gradcam++}
Aditya Chattopadhay, Anirban Sarkar, Prantik Howlader, and Vineeth~N Balasubramanian.
\newblock Grad-cam++: Generalized gradient-based visual explanations for deep convolutional networks.
\newblock In \emph{IEEE Winter Conference on Applications of Computer Vision}, pages 839--847, 2018.

\bibitem[Jiang et~al.(2021)Jiang, Zhang, Hou, Cheng, and Wei]{jiang2021layercam}
Peng-Tao Jiang, Chang-Bin Zhang, Qibin Hou, Ming-Ming Cheng, and Yunchao Wei.
\newblock Layercam: Exploring hierarchical class activation maps for localization.
\newblock \emph{IEEE Transactions on Image Processing}, 30:\penalty0 5875--5888, 2021.

\bibitem[Deng et~al.(2009)Deng, Dong, Socher, Li, Li, and Fei-Fei]{deng2009imagenet}
Jia Deng, Wei Dong, Richard Socher, Li-Jia Li, Kai Li, and Li~Fei-Fei.
\newblock Imagenet: A large-scale hierarchical image database.
\newblock In \emph{IEEE Computer Society Conference on Computer Vision and Pattern Recognition}, pages 248--255, 2009.

\bibitem[Van~Horn et~al.(2018)Van~Horn, Mac~Aodha, Song, Cui, Sun, Shepard, Adam, Perona, and Belongie]{van2018inaturalist}
Grant Van~Horn, Oisin Mac~Aodha, Yang Song, Yin Cui, Chen Sun, Alex Shepard, Hartwig Adam, Pietro Perona, and Serge Belongie.
\newblock The inaturalist species classification and detection dataset.
\newblock In \emph{Proceedings of the IEEE Conference on Computer Vision and Pattern Recognition}, pages 8769--8778, 2018.

\bibitem[Shrikumar et~al.(2016)Shrikumar, Greenside, Shcherbina, and Kundaje]{gradient_input}
Avanti Shrikumar, Peyton Greenside, Anna Shcherbina, and Anshul Kundaje.
\newblock Not just a black box: Learning important features through propagating activation differences.
\newblock \emph{arXiv preprint arXiv:1605.01713}, 2016.

\bibitem[Sundararajan et~al.(2017)Sundararajan, Taly, and Yan]{integrated_gradients}
Mukund Sundararajan, Ankur Taly, and Qiqi Yan.
\newblock Axiomatic attribution for deep networks.
\newblock In \emph{International Conference on Machine Learning}, pages 3319--3328, 2017.

\bibitem[Deng et~al.(2023)Deng, Zou, Du, Chen, Feng, Yang, Li, and Zhang]{unified_attribution}
Huiqi Deng, Na~Zou, Mengnan Du, Weifu Chen, Guocan Feng, Ziwei Yang, Zheyang Li, and Quanshi Zhang.
\newblock Understanding and unifying fourteen attribution methods with taylor interactions.
\newblock \emph{arXiv preprint arXiv:2303.01506}, 2023.

\bibitem[Khakzar et~al.(2022)Khakzar, Khorsandi, Nobahari, and Navab]{zero_attribution}
Ashkan Khakzar, Pedram Khorsandi, Rozhin Nobahari, and Nassir Navab.
\newblock Do explanations explain? model knows best.
\newblock In \emph{Proceedings of the IEEE/CVF Conference on Computer Vision and Pattern Recognition}, pages 10244--10253, 2022.

\bibitem[Huang and Li(2021)]{benchmark_imagenet}
Rui Huang and Yixuan Li.
\newblock Mos: Towards scaling out-of-distribution detection for large semantic space.
\newblock In \emph{Proceedings of the IEEE/CVF Conference on Computer Vision and Pattern Recognition}, pages 8710--8719, 2021.

\bibitem[Sun et~al.(2022)Sun, Ming, Zhu, and Li]{KNN}
Yiyou Sun, Yifei Ming, Xiaojin Zhu, and Yixuan Li.
\newblock Out-of-distribution detection with deep nearest neighbors.
\newblock In \emph{International Conference on Machine Learning}, pages 20827--20840, 2022.

\bibitem[Djurisic et~al.(2023)Djurisic, Bozanic, Ashok, and Liu]{ASH}
Andrija Djurisic, Nebojsa Bozanic, Arjun Ashok, and Rosanne Liu.
\newblock Extremely simple activation shaping for out-of-distribution detection.
\newblock In \emph{International Conference on Learning Representations}, 2023.

\bibitem[He et~al.(2016)He, Zhang, Ren, and Sun]{ResNet}
Kaiming He, Xiangyu Zhang, Shaoqing Ren, and Jian Sun.
\newblock Deep residual learning for image recognition.
\newblock In \emph{Proceedings of the IEEE Conference on Computer Vision and Pattern Recognition}, pages 770--778, 2016.

\bibitem[Zagoruyko and Komodakis(2016)]{wideResNet}
Sergey Zagoruyko and Nikos Komodakis.
\newblock Wide residual networks.
\newblock \emph{arXiv preprint arXiv:1605.07146}, 2016.

\bibitem[Xiao et~al.(2010)Xiao, Hays, Ehinger, Oliva, and Torralba]{xiao2010sun}
Jianxiong Xiao, James Hays, Krista~A Ehinger, Aude Oliva, and Antonio Torralba.
\newblock Sun database: Large-scale scene recognition from abbey to zoo.
\newblock In \emph{2010 IEEE Computer Society Conference on Computer Vision and Pattern Recognition}, pages 3485--3492, 2010.

\bibitem[Zhou et~al.(2017)Zhou, Lapedriza, Khosla, Oliva, and Torralba]{zhou2017places}
Bolei Zhou, Agata Lapedriza, Aditya Khosla, Aude Oliva, and Antonio Torralba.
\newblock Places: A 10 million image database for scene recognition.
\newblock \emph{IEEE Transactions on Pattern Analysis and Machine Intelligence}, 40\penalty0 (6):\penalty0 1452--1464, 2017.

\bibitem[Cimpoi et~al.(2014)Cimpoi, Maji, Kokkinos, Mohamed, and Vedaldi]{cimpoi2014describing}
Mircea Cimpoi, Subhransu Maji, Iasonas Kokkinos, Sammy Mohamed, and Andrea Vedaldi.
\newblock Describing textures in the wild.
\newblock In \emph{Proceedings of the IEEE Conference on Computer Vision and Pattern Recognition}, pages 3606--3613, 2014.

\bibitem[Netzer et~al.(2011)Netzer, Wang, Coates, Bissacco, Wu, and Ng]{SVHN}
Yuval Netzer, Tao Wang, Adam Coates, Alessandro Bissacco, Bo~Wu, and Andrew~Y Ng.
\newblock Reading digits in natural images with unsupervised feature learning.
\newblock 2011.

\bibitem[Yu et~al.(2015)Yu, Seff, Zhang, Song, Funkhouser, and Xiao]{LSUN}
Fisher Yu, Ari Seff, Yinda Zhang, Shuran Song, Thomas Funkhouser, and Jianxiong Xiao.
\newblock Lsun: Construction of a large-scale image dataset using deep learning with humans in the loop.
\newblock \emph{arXiv preprint arXiv:1506.03365}, 2015.

\bibitem[Kolesnikov et~al.(2020)Kolesnikov, Beyer, Zhai, Puigcerver, Yung, Gelly, and Houlsby]{resnetv2}
Alexander Kolesnikov, Lucas Beyer, Xiaohua Zhai, Joan Puigcerver, Jessica Yung, Sylvain Gelly, and Neil Houlsby.
\newblock Big transfer (bit): General visual representation learning.
\newblock In \emph{ECCV}, pages 491--507. Springer, 2020.

\bibitem[Guidotti et~al.(2018)Guidotti, Monreale, Ruggieri, Turini, Giannotti, and Pedreschi]{guidotti2018survey}
Riccardo Guidotti, Anna Monreale, Salvatore Ruggieri, Franco Turini, Fosca Giannotti, and Dino Pedreschi.
\newblock A survey of methods for explaining black box models.
\newblock \emph{ACM Computing Surveys}, 51\penalty0 (5):\penalty0 1--42, 2018.

\bibitem[Nguyen et~al.(2015)Nguyen, Yosinski, and Clune]{networks_fooled}
Anh Nguyen, Jason Yosinski, and Jeff Clune.
\newblock Deep neural networks are easily fooled: High confidence predictions for unrecognizable images.
\newblock In \emph{Proceedings of the IEEE Conference on Computer Vision and Pattern Recognition}, pages 427--436, 2015.

\bibitem[Hein et~al.(2019)Hein, Andriushchenko, and Bitterwolf]{OOD_overconfidence_hein2019relu}
Matthias Hein, Maksym Andriushchenko, and Julian Bitterwolf.
\newblock Why relu networks yield high-confidence predictions far away from the training data and how to mitigate the problem.
\newblock In \emph{Proceedings of the IEEE/CVF Conference on Computer Vision and Pattern Recognition}, pages 41--50, 2019.

\bibitem[Dosovitskiy et~al.(2020)Dosovitskiy, Beyer, Kolesnikov, Weissenborn, Zhai, Unterthiner, Dehghani, Minderer, Heigold, Gelly, et~al.]{vit}
Alexey Dosovitskiy, Lucas Beyer, Alexander Kolesnikov, Dirk Weissenborn, Xiaohua Zhai, Thomas Unterthiner, Mostafa Dehghani, Matthias Minderer, Georg Heigold, Sylvain Gelly, et~al.
\newblock An image is worth 16x16 words: Transformers for image recognition at scale.
\newblock \emph{International Conference on Learning Representations}, 2020.

\end{thebibliography}

\end{document}